%% file: emnlp2022.tex
\pdfoutput=1
\documentclass[11pt]{article}

\usepackage{EMNLP2022}

\usepackage{times}
\usepackage{latexsym}

\usepackage[T1]{fontenc}
\usepackage{xspace}
\usepackage{siunitx}
\usepackage{xcolor}
\usepackage{makecell}
\usepackage{booktabs}
\usepackage{hyperref}
\usepackage{mathtools}
\usepackage{amsfonts}
\usepackage{amsmath,amssymb}
\usepackage{nicefrac}       % compact symbols for 1/2, etc.
\usepackage{empheq}
\usepackage[utf8]{inputenc}
\usepackage{amsmath}
\usepackage{graphicx}
\usepackage{bm}
\usepackage[shortlabels]{enumitem}
\usepackage{cleveref}
\Crefformat{section}{\S#2#1#3}
\Crefformat{subsection}{\S#2#1#3}
\Crefformat{subsubsection}{\S#2#1#3}
\Crefrangeformat{section}{\S\S#3#1#4 to~#5#2#6}
\Crefmultiformat{section}{\S\S#2#1#3}{ and~#2#1#3}{, #2#1#3}{ and~#2#1#3}
\usepackage{refstyle}
\Crefformat{figure}{#2Fig.~#1#3}
\Crefmultiformat{figure}{Figs.~#2#1#3}{ and~#2#1#3}{, #2#1#3}{ and~#2#1#3}
\Crefformat{table}{#2Tab.~#1#3}
\Crefmultiformat{table}{Tabs.~#2#1#3}{ and~#2#1#3}{, #2#1#3}{ and~#2#1#3}
\Crefformat{appendix}{#2Appx.~\S#1#3}
\Crefformat{algorithm}{Alg.~#2#1#3}
\Crefformat{equation}{#2Eq.~#1#3}
\usepackage{colortbl}
\newcommand{\CC}[1]{\cellcolor{blue!#1}}
\newcommand{\RED}[1]{\cellcolor{red!#1}}

\newcommand{\temprel}{\textsc{TempRel}\xspace}

\newcommand{\zb}{\mathbf{z}}
\newcommand{\yb}{\mathbf{y}}
\newcommand{\stitle}[1]{\vspace{1ex}\noindent{\bf #1}}

\DeclareMathOperator{\acc}{acc}
\DeclareMathOperator{\conf}{conf}
\newcommand{\scrL}{\mathcal{L}}
\newcommand{\xb}{\bar{X}}

\usepackage{microtype}

\title{\emph{Extracting or Guessing?} Improving Faithfulness of Event Temporal Relation Extraction}

\author{\makecell{Haoyu Wang$^1$, Hongming Zhang$^2$, Yuqian Deng$^1$, \\Jacob R. Gardner$^1$,   Dan Roth$^1$ \& Muhao Chen$^{3}$}\\
$^1$Department of Computer and Information Science, UPenn\\
$^2$Tencent AI Lab, Seattle\\
$^3$Department of Computer Science, USC\\
\texttt{\{why16gzl, yuqiand, jacobrg, danroth\}@seas.upenn.edu};\\ \texttt{hongmzhang@tencent.com};
\texttt{muhaoche@usc.edu}\\
}

\begin{document}
\maketitle
\begin{abstract}
In this paper, we seek to improve the faithfulness of \temprel extraction models from two perspectives.
The \textit{first} perspective is to extract genuinely based on contextual description. 
To achieve this, we propose to conduct counterfactual analysis to attenuate the effects of two significant types of training biases: the \textit{event trigger bias} and the \textit{frequent label bias}.
We also add tense information into event  representations to explicitly place an emphasis on the contextual description.
The \textit{second} perspective is to provide proper uncertainty estimation and abstain from extraction when no relation is described in the text.
%To know when to abstain, the model needs to have a good estimate of the true correctness likelihood. 
By parameterization of Dirichlet Prior over the model-predicted categorical distribution, we improve the model estimates of the correctness likelihood and make \temprel predictions more selective.
We also employ temperature scaling to recalibrate the model confidence measure after bias mitigation.
Through experimental analysis on MATRES, MATRES-DS, and TDDiscourse, we demonstrate that our model extracts \temprel and timelines more faithfully compared to SOTA methods, especially under distribution shifts.
\end{abstract}

\section{Introduction}
\label{sec:intro}
Event temporal relation (\temprel) extraction is an essential step towards understanding narrative text, such as stories, novels, news, and guideline articles. 
With a robust temporal relation extractor, one can easily construct a storyline from text and capture the trend of temporally connected event mentions. 
\temprel extraction is also broadly beneficial to various downstream tasks including clinical narrative processing \cite{jindal-roth-2013-using, bethard-etal-2016-semeval}, question answering \cite{llorens-etal-2015-semeval, meng-etal-2017-temporal, stricker-2021-question-answering}, and schema induction \cite{chambers-jurafsky-2009-unsupervised, wen-etal-2021-resin, li-etal-2021-future}.

Most existing \temprel extraction models are developed with data-driven machine learning approaches, %leveraging different techniques ranging from advanced training paradigms and data augmentation \cite{ning-etal-2017-structured, ballesteros-etal-2020-severing, mathur-etal-2021-timers, tan-etal-2021-extracting, trong2022selecting} to incidental supervision \cite{ning-etal-2018-improving, han-etal-2019-joint, wang-etal-2020-joint, zhao-etal-2021-effective, zhou-etal-2021-temporal}.
for which recent studies also incorporate advanced learning and inference techniques such as structured prediction \cite{ning-etal-2017-structured,ning-etal-2018-improving,han-etal-2019-joint,wang-etal-2020-joint,tan-etal-2021-extracting}, graph representation \cite{mathur-etal-2021-timers, qiang-naacl-2022}, data augmentation \cite{ballesteros-etal-2020-severing,trong2022selecting}, and indirect supervision \cite{zhao-etal-2021-effective, zhou-etal-2021-temporal}.
These models are prevalently built upon pretrained language models (PLMs) and fine-tuned on a small set of annotated documents, e.g., TimeBank-Dense \cite{cassidy-etal-2014-annotation}, MATRES \cite{ning-etal-2018-multi}, and TDDiscourse \cite{naik-etal-2019-tddiscourse}.

% Common probelm: *Faithness* in 2 perspective: 
% 1. extract genuinely based on what is described in context (but not giving trivial guess from event names)
% 2. abstain to extract when no relation is described. (selective)

% Commonality in prior methods: built on PLMs, training with (small #) annotated documents.
% Common problems: 
% A. Learning bias
% 1. prior knowledge in PLMs may be conflict with contexts in inference.
% 2. Training data bias: 
% 2.1. biases of event triggers/names: distribution of appeared triggers are skewed.
% 2.2. label/relation biases
% -> models may capture spurious correlation between names and relations, leading to shortcutted prediction.
% B. Not handling NoRel or Vague cases
% Those are exception cases in the IE task. However, such exceptions are never close to exhaustive, or even not given.
% -> models need to learn to *selectively* predict
\begin{figure}[t]
\begin{minipage}{\linewidth}
\noindent
\fbox{%
    \parbox{0.95\linewidth}{
    \fontsize{11pt}{13pt}\selectfont
    A) I went to \textcolor{blue}{\textsc{$e_1$:see}} the doctor. However, I was more seriously \textcolor{blue}{\textsc{$e_2$:sick}}. $\Longrightarrow$ \textcolor{blue}{\textsc{$e_1$}} \textsc{After} \textcolor{blue}{\textsc{$e_2$}} \\ % prior knowledge
    
    B) Microsoft said it has \textcolor{blue}{\textsc{$e_3$:identified}} three companies for \textit{the China program} to run through June. The company also \textcolor{blue}{\textsc{$e_4$:gives}} each participating startup in \textit{the Seattle program} \$20,000 to create software. $\Longrightarrow$ \textcolor{blue}{\textsc{$e_3$}} \textsc{Before} \textcolor{blue}{\textsc{$e_4$}}% statistical bias from training data \\
    }
}
\vspace{-0.5em}
    \caption{Examples of unfaithful extractions. \textsc{Before} and \textsc{After} that follow the arrows denote the extracted \temprel's from the sentences by \cite{zhou-etal-2021-temporal}.}\label{fig:example}
    \vspace{-0.5em}
\end{minipage}
\end{figure}

Though these %state-of-the-art
recent approaches have achieved promising evaluation results on benchmarks, 
%\muhao{at here we can directly point out that whether the models are making faithful extractions is an unexplored issue. Then 1. define faithfulness: 1. genuinely extract what is exactly described in the context; 2. know when there is no relation to extract.
%Then the faithfulness issue is attributed to the following biases from prior knowledge or biased training.}
whether they provide \emph{faithful} extraction is an unexplored problem.
The \emph{faithfulness} of a relation extraction system is not simply about how much accuracy a system can offer.
Instead, a faithful extractor should concern the validity and reliability of its extraction process. 
Specifically, when there is a \temprel to extract, a faithful extractor should genuinely obtain what is described in the context but not give \emph{trivial guesses} from surface names of events or most frequent labels.
Besides, when there is no relation described in the context, the system should selectively abstain from prediction.
%\muhao{+ the reality is that biases from pretrained prior knowledge and training biases lead to unfaithful extractions. (can give two examples of extracts to illustrate them.)}

We observe that in recent models, biases from prior knowledge in PLMs and statistically skewed training data often lead to unfaithful extractions (see \Cref{fig:example}).
Example A thereof exhibits a case where the model adheres to the prior knowledge where people usually \textbf{see the doctor} \emph{after} \textbf{getting sick}, but in this context %\textbf{\textit{got more sick}} in the context is describing a semantically different event from the commonly mentioned one and hence the correct \temprel here is \textsc{Before}.
\textbf{getting sick} is obviously a consequent of \textbf{seeing the doctor}. 
In Example B, \textsc{Before} is extracted due to statistical biases learned from training data that \textsc{Before} is not only the most frequent \temprel between \textbf{\textit{identify}} and \textbf{\textit{give}},
but is also the most frequent \temprel between the first and second event in narrative order \cite{gee1984empirical}.
However, with a closer inspection, it can be noticed that the two events in Example B are involved in different programs, one in \textit{the China program}, the other in \textit{the Seattle program}.
Therefore, the system should abstain from prediction and give \textsc{Vague} as output.

%There are mainly two kinds of bias: First, given a skewed distribution of appeared triggers in the learning corpus, the models can easily learn from event names to predict the \temprel between them. Second, the models may learn to make shortcutted predictions with imbalanced training labels.
%Moreover, they are not handling \textsc{Vague} cases well since \textsc{Vague} relations are exceptional in the \temprel extraction task and such exceptions are never close to exhaustive, and sometimes are even not given in the training corpus \cite{naik-etal-2019-tddiscourse}.

In this paper, we seek to improve the faithfulness of \temprel extraction models from two perspectives.
The \textit{first} perspective is to guide the model to genuinely extract the described \temprel based on a relation-mentioning context. % instead of providing trivial guesses under training biases' sway.
%To acheive this goal, we propose two techniques to adjust the event representation and relation inference. 
To achieve this goal, we conduct counterfactual analysis \cite{niu2021counterfactual} 
to capture and attenuate the effects of two typical types of training biases: \textit{event bias} caused by treating event trigger names as shortcuts for \temprel prediction,
and \textit{label bias} that causes the model prediction to lean towards more frequent training labels.
We also propose to affix tense information to event mentions to explicitly place an emphasis on the contextual description.
%to obtain event representations 

%The two significant types of training biases, event surface names and frequent label biases, are mitigated by subtracting weighted probability distribution given partial or zero input information.

%\muhao{selective prediction}
The \textit{second} perspective is to %abstain from extraction when no relation
%realize selective prediction \cite{geifman2017selective},
teach the model to abstain from extraction when no relation
is described in the text. 
To know when to abstain, the models need to have a good estimate of the correctness likelihood. 
By incorporating Dirichlet Prior \cite{malinin2018, malinin2019reverse} in the training phase of current \temprel extraction models, we improve the predictive uncertainty estimation of the models and make the \temprel predictions more selective.
Furthermore, since the counterfactual analysis component (from the \textit{first} perspective) %the models' output probability distribution over \temprel labels is shifted towards unbiased directions.
may shift the model-predicted categorical  distribution,
we also employ temperature scaling \cite{pmlr-v70-guo17a} in inference to allow for recalibrated confidence measure of the model.
%To tune the models back to providing calibrated confidence measure, we employ temperature scaling \cite{pmlr-v70-guo17a} and find the optimal temperature using validation sets of the benchmarks.

The technical contributions of our work are two-folds.
First, to the best of our knowledge, this is the first
study on the faithfulness issue of event-centric information extraction.
Evidently, the development of a faithful \temprel extraction system contributes to more robust and reliable machine comprehension of events and narratives.
Second, we propose training and inference techniques that can be easily plugged into existing neural \temprel extractors and effectively improve model faithfulness by mitigating prediction shortcuts and enhancing the capability of selective prediction.
%We evaluate our method regarding both preciseness and faithfulness on two tasks: \temprel extraction and timeline construction \cite{do-etal-2012-joint}.

Our contributions are verified with \temprel extraction experiments conducted on MATRES \cite{ning-etal-2018-multi}, %MATRES-DS \todo{needs explanation?}, and 
TDDiscourse \cite{naik-etal-2019-tddiscourse} %for as our training data and use
and distribution-shifted version of MATRES (MATRES-DS).
%where we create three different evaluation settings that concern in-distribution data and data under distribution shifts. 
Particularly, we evaluate on how precise and selective our \temprel extraction method is on in-distribution data, and how well it generalizes under distribution shift.
Experimental results demonstrate that the techniques explored within the two aforementioned perspectives bring about promising results in improving faithfulness of current models.
In addition, we also apply our method to the task of timeline construction \cite{do-etal-2012-joint},
showing that faithful \temprel extraction greatly benefits the accurate construction of timelines. %effective understanding of news events.

\section{Related Work}

%We discuss three lines of relevent research.

\stitle{Event \temprel Extraction.}
Recent event \temprel extraction approaches are mainly built on PLMs to obtain representations of event mentions and are improved with various learning and inference methodologies. 
To improve the quality of event representations, \citet{mathur-etal-2021-timers} embrace rhetorical discourse features and temporal arguments; \citet{trong2022selecting} select optimal context sentences via reinforcement learning to achieve SOTA performances; while \citet{liu2021discourse,mathur-etal-2021-timers,qiang-naacl-2022} employ graph neural networks to avoid complex feature engineering.
From the learning perspective, \citet{ning-etal-2018-joint}, \citet{ballesteros-etal-2020-severing}, and \citet{wang-etal-2020-joint} enrich the models with %useful information
auxiliary training tasks to provide complementary supervision signals, while \citet{ning-etal-2018-improving}, \citet{zhao-etal-2021-effective} and \citet{zhou-etal-2021-temporal} bring into play distant supervision from heuristic cues and patterns.
Nevertheless, recent data-driven models risk amplifying bias by exacerbating biases present in the %training set
pretraining and task training data when making predictions \cite{zhao-etal-2017-men}. 
To rectify the models' biases towards prior knowledge in PLMs and shortcuts learned from biased training %and distant 
examples, our work proposes several training and inference techniques, seeking to improve the faithfulness of neural \temprel extractors as described in \Cref{sec:intro}.

\stitle{Bias Mitigation in NLP.} 
Methods for %bias mitigation
mitigating prediction biases can be categorized as retraining and inference \cite{sun-etal-2019-mitigating}.
Retraining methods address the bias in early stages or at its source.
For instance, \citet{zhang-etal-2017-position} masks the entities with special tokens to prevent relation extraction models from learning shortcuts from entity names, whereas several works conduct data augmentation \cite{park-etal-2018-reducing,alzantot-etal-2018-generating,Jin_Jin_Zhou_Szolovits_2020,wu-etal-2022-generating} or sample reweighting techniques \cite{lin2017focal,liu2021just} to reduce biases in training. %respectively reduce gender bias by data augmentation and fine-tuning on unbiased data. 
However, masking would result in the loss of semantic information and performance degradation, and it is costly to manipulate data or find proper unbiased data in temporal reasoning.
Directly debiasing the training process may also hinder the model generalization on out-of-distribution (OOD) data \cite{wang2022should}.
Therefore, inspired by several recent studies on debiasing text classification or entity-centric information extraction \cite{qian-etal-2021-counterfactual, nan-etal-2021-uncovering}, our work adopts counterfactual inference to %focus on contextual reasoning via causal graphs, 
measure and control prediction biases based on automatically generated counterfactual examples. 
%as is studied by . 

\stitle{Selective Prediction.}
Neural models have become increasingly accurate with the advances of deep learning.
In the meantime, however, they should also indicate when their predictions are likely to be inaccurate in real-world scenarios.
A series of recent studies have focused on resolving model miscalibration by measuring how closely the model confidences match empirical likelihoods.
Among them, computationally expensive Bayesian \cite{gal2016dropout, Kueppers_2021_IV} and non-Bayesian ensemble \cite{NIPS2017_9ef2ed4b, Beluch_2018_CVPR} methods have been adopted to yield high quality predictive uncertainty estimates.
Other methods have been proposed to use uncertainty reflected from model parameters to assess the confidence, including sharpness \cite{kuleshov2018accurate} and softmax response \cite{hendrycks2017baseline,xin-etal-2021-art}.
Another class of methods adjust the models' output probability distribution by altering loss function in training via label smoothing \cite{szegedy2016rethinking} and Dirichlet Prior \cite{malinin2018, malinin2019reverse}.
Besides, temperature scaling \cite{pmlr-v70-guo17a} %, desai-durrett-2020-calibration, dan-roth-2021-effects-transformer, NEURIPS2021_8420d359} 
also serves as a simple yet effective post-hoc calibration technique. 
In this paper, we model \temprel's with Dirichlet Prior in learning, and 
%treat \textsc{Vague} relations annotated in \temprel datasets as out-of-distribution instances and separately model them via Dirichlet Prior distributional uncertainty in the training phase, following \cite{NEURIPS2018_3ea2db50}. 
during inference we employ temperature scaling to recalibrate confidence measure of the model after bias mitigation.

\section{Preliminaries}
\label{sec:prelim}
A document $D$ is represented as a sequence of tokens $D = [w_1, \cdots, e_1, \cdots, e_2, \cdots, w_m]$,
where some tokens belong to the set of annotated event triggers, i.e., $\mathcal{E}_D = \{e_1, e_2, \cdots, e_n\}$,
and the rest are other lexemes.
For a pair of events $(e_i, e_j)$, the task of \temprel extraction is to predict a relation $r$ from $\mathcal{R} \cup \{\textsc{Vague}\}$, where $\mathcal{R}$ denotes the set of \temprel's.
An event pair is labeled \textsc{Vague} if the text does not express any determinable relation that belongs to $\mathcal{R}$.
%Let $\zb^{(i, j)}$ be the \emph{logits vector} produced before the softmax layer of the neural model and $\yb^{(i, j)}$ denote the \emph{model-predicted categorical distribution} over $\mathcal{R}$.
Let $\yb^{(i, j)}$ denote the \emph{model-predicted categorical distribution} over $\mathcal{R}$.
%We allow the model to abstain from extraction on low-confidence examples and output \textsc{Vague}.
%In this work, we would like our model to not only genuinely extract \temprel from contextual descriptions of events, but 

In order to provide a confidence estimate $\yb$ that is as close as possible to the true probability, we first describe three separate factors \cite{malinin2018} that attribute to the predictive uncertainty for an AI system, namely epistemic uncertainty, aleatoric uncertainty, and distributional uncertainty.
Epistemic uncertainty refers to the degree of uncertainty in estimating model parameters based on training data, whereas aleatoric uncertainty results from data's innate complexity. Distributional uncertainty arises when the model cannot make accurate predictions due to the lack of familiarity with the test data.

We argue that the way of handling \textsc{Vague} relations in existing \temprel extractors is problematic since they typically merge \textsc{Vague} into $\mathcal{R}$.
In fact, \textsc{Vague} relations are complicated exception cases in the IE task, yet the annotation of such exceptions are never close to exhaustive in benchmarks, or even not given \cite{naik-etal-2019-tddiscourse}.
In this work, we consider \textsc{Vague} relations as a source of \textbf{distributional uncertainty} and separately model them.
Details are introduced in \Cref{sec:uncertainty_esti}.

\section{Methods}
In this section, we first present how we obtain event representations and categorical distribution $\yb$ in a local classifier for \temprel (\Cref{sec:event_rep}). Then we introduce proposed learning and inference techniques to improve model faithfulness from the perspectives of selective prediction (\Cref{sec:uncertainty_esti}) and prediction bias mitigation  (\Cref{sec:counterfactual_analysis}), % as well as temperature scaling that is necessary to combine these two techniques (\Cref{sec:ts}). 
before we combine these two techniques with temperature scaling and introduce the OOD detection method in \Cref{sec:ts}.
%We conclude this section by illustrating the overall architecture of our \temprel extractor and the algorithm developed for timeline construction in \Cref{sec:model_arch} and \Cref{sec:timeline} respectively.

\subsection{Local Classifier}
\label{sec:event_rep}
Given that the context around an event pair $(e_i, e_j)$\footnote{For event pair $(e_i, e_j)$ we assume $e_i$ is the first event in narrative order if $i < j$.} has linguistic signals and temporal cues that are beneficial to \temprel prediction, the context of $(e_i, e_j)$ considered in our model starts from the sentence before $e_i$ and ends at the sentence after $e_j$.
Inspired by \citet{ZC21} for improving entity representation by prepending entity type information to entity mention spans, we add %event temporal-related information
%temporal aspect
tense information of events into event trigger representations in this work.
Accordingly, we enclose $e_i$ and $e_j$ with ``$@$'' and ``\#'' respectively\footnote{Note that similar to  \emph{typed entity marker (punct)} by \citet{ZC21}, such enclosing has the benefit of highlighting mention spans without introducing new special tokens.} and prepend their tense information to their spans with ``$*$'' and ``$\wedge$''.
%The modified text is in the form of ``@ * \textsc{\textit{$t_i$}} * \textsc{$e_i$} @ ... \# $\wedge$ \textsc{\textit{$t_j$}} $\wedge$ \textsc{$e_j$} \# '', where \textsc{\textit{$t_i$}} and \textsc{\textit{$t_j$}} are the tense of $e_i$ and $e_j$. 
% acquired from an off-the-shelf tense identifier\footnote{\url{https://tense-sense-identifier.herokuapp.com/}}.
We provide a detailed example for affixing tense information in \Cref{sec:tense_eg}. %\Cref{fig:affix}.

To characterize event pair $(e_i, e_j)$, we obtain the two events' contextual representations and attention heads from PLMs.
The classifer is trained to uncover the context that is critical to both events by multiplying their attentions before we send the concatenation of token embeddings and attention multiplication to a multi-layer perceptron (MLP) with $|\mathcal{R}|$ outputs.
In this fashion, we obtain the $|\mathcal{R}|$-dimensional logits vector $\zb^{(i, j)}$ and categorical distribution $\yb^{(i, j)}$, where
the probability of a label $r \in \mathcal{R}$ is given by the softmax function $\sigma(\cdot)$:
\begin{equation}\label{eqn:softmax}
    y_r = \sigma(\zb)_r = \frac{ e^{z_r}}{\sum_{k=1}^{|\mathcal{R}|} e^{z_k}} \ .
\end{equation}

\subsection{Parameterization of Dirichlet Prior}
\label{sec:uncertainty_esti}
\begin{figure}[t]
    \centering
    \includegraphics[width=0.81\linewidth]{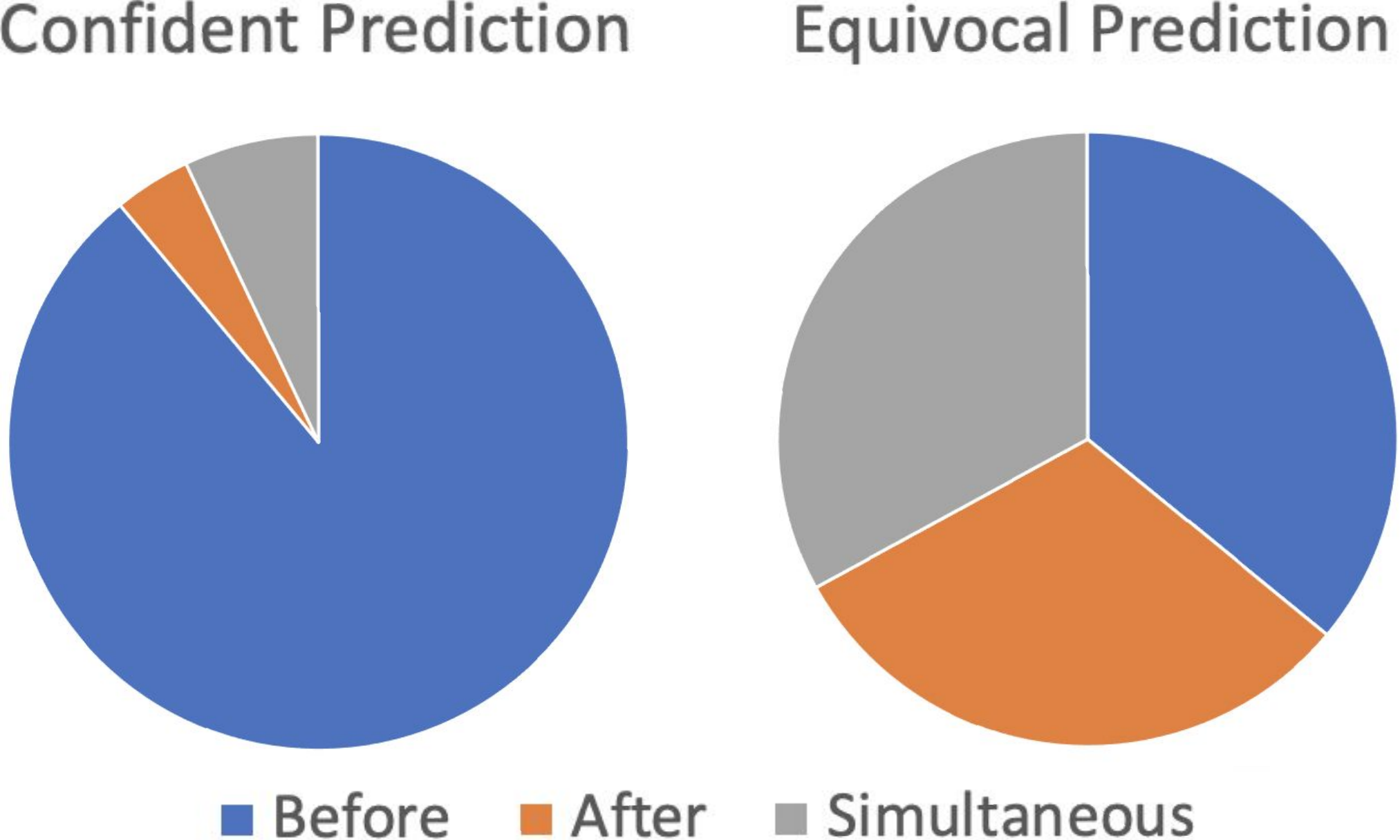}
     \vspace{-0.5em}
    \caption{An illustration for desired behaviors of model predicted categorical distribution.}
    \label{fig:behavior}
     \vspace{-0.5em}
\end{figure}

As discussed in preliminaries (\Cref{sec:prelim}), \textsc{Vague} %relations are
corresponds to complicated exception cases in inference.
We model them as out-of-distribution (OOD) cases which are different from in-distribution (ID) data describing the relations in $\mathcal{R}$. %that are in-distribution (ID).
The goal of providing high-quality confidence estimate $\yb$ requires the model to yield a sharp predicted distribution centered on one of the labels in $\mathcal{R}$ when it is confident and yield a flat distribution over $\mathcal{R}$ for OOD inputs, as is shown in \Cref{fig:behavior}.
To achieve this goal, we explicitly \textbf{parameterize a prior distribution} over categorical distributions.
Because of the tractable analytic properties\footnote{$\Gamma(\cdot)$ in \Cref{eqn:dirichlet} denotes the gamma function.} of Dirichlet distribution (\Cref{eqn:dirichlet}), we choose to parameterize a \emph{sharp} and a \emph{flat} Dirichlet prior over the model-predicted categorical distribution for ID and OOD inputs, respectively. 
The Dirichlet distribution is parameterized by its concentration parameters $\bm{\alpha}$, where $\alpha_0$ is the precision of Dirichlet distribution. Higher values of $\alpha_0$ lead to sharper, more confident predicted distributions. 
\begin{equation}\label{eqn:dirichlet}
\begin{aligned}
{\tt Dir}(\bm{\yb};\bm{\alpha}) = \frac{\Gamma(\alpha_0)}{\prod_{k=1}^{|\mathcal{R}|}\Gamma(\alpha_k)}\prod_{k=1}^{|\mathcal{R}|} y_k^{\alpha_k -1} , \\
\alpha_k >0 ,\ \alpha_0 = \sum_{k=1}^{|\mathcal{R}|} \alpha_k \ .
\end{aligned}
\end{equation} 

\begin{figure*}[t]
    \centering
    \includegraphics[width=\linewidth]{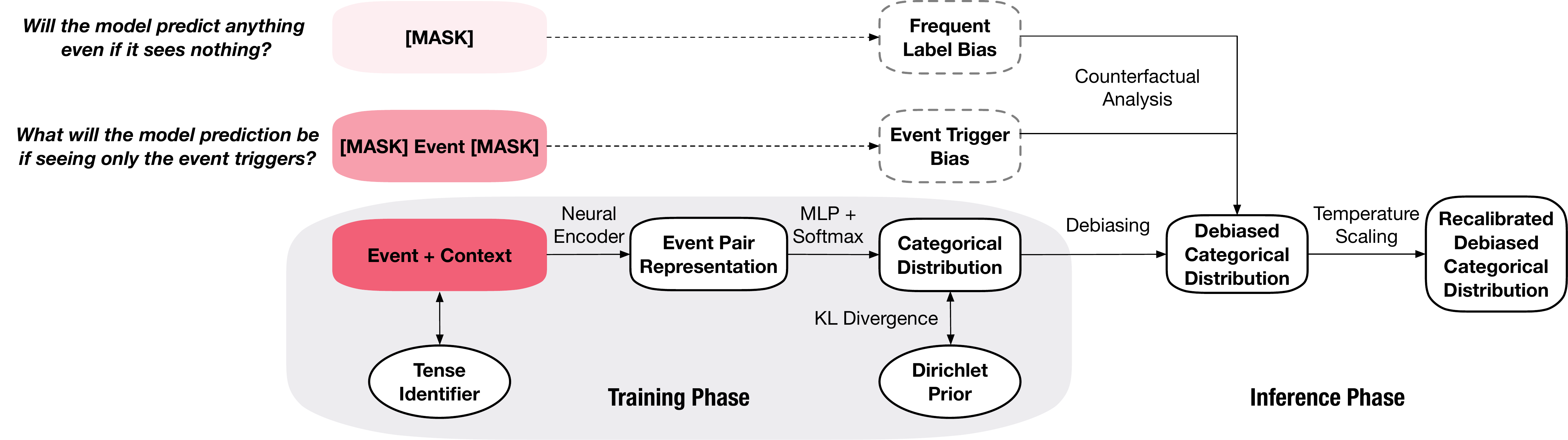}
    \vspace{-1em}
    \caption{An overview of our approach to improving model faithfulness. In the training phase we obtain the model-predicted categorical distribution $\yb$ with a neural encoder and parameterize a Dirichlet Prior over $\yb$. And then we conduct counterfactual analysis to distill and mitigate biases during inference before leveraging temperature scaling to obtain recalibrated and debiased $\yb$.}
    \label{fig:model}
    \vspace{-0.5em}
\end{figure*}

To attain the aforementioned behaviors, on ID data the model is trained to minimize the KL divergence between a sharp Dirichlet distribution and the model-predicted categorical distribution:
\begin{equation}
\label{eqn:id_loss}
\mathcal{L}_{ID} = \displaystyle \mathop{\mathbb{E}_{{\tt p_{ID}}(\bm{x})}}[KL[{\tt p}(\yb) \parallel {\tt Dir}(\yb;\bm{\alpha}_{s})]],
\end{equation}
where ${\tt p_{ID}}(\bm{x})$ denotes ID data and $\bm{\alpha}_{s}$ denotes concentration parameters of the sharp Dirichlet distribution.
On OOD data, the model minimizes the KL divergence between a flat Dirichlet distribution and the model-predicted categorical distribution:
\begin{equation}
\label{eqn:ood_loss}
\mathcal{L}_{OOD} = \mathbb{E}_{{\tt p_{OOD}}(\bm{x})}[KL[{\tt p}(\yb) \parallel {\tt Dir}(\yb;\bm{\alpha}_{f})]],
\end{equation}
where ${\tt p_{OOD}}(\bm{x})$ denotes OOD data and $\bm{\alpha}_{f}$ denotes the concentration parameters of the flat Dirichlet distribution.
And the total loss of the model is 
\begin{equation}
    \label{eqn:loss}
    \mathcal{L} = \lambda_1 \mathcal{L}_{ID} + \lambda_2 \mathcal{L}_{OOD},
\end{equation}
where the $\lambda$'s are hyperparameters to balance the influence of each loss.
With the parameterization of Dirichlet prior, the learning process seeks to partly enhance the model's faithfulness by outputting confident estimates when it encounters ID inputs, and outputting equivocal estimates when the context does not express any \temprel in the meantime.

\subsection{Counterfactual Analysis}\label{sec:counterfactual_analysis}
After looking into the selective prediction perspective of faithfulness, we now address the other perspective: to mitigate biases %learned 
from %the training data
pre-trained knowledge and the task training data during the inference stage.
Given that we have observed two types of biases in existing models, namely the event trigger bias and the frequent label bias, we ask the following questions:
\begin{itemize}[leftmargin=*]
\setlength\itemsep{-0.3em}
    \item \textit{What will the model prediction be if seeing the full context?}
    \item \textit{What will the model prediction be if seeing only the event tiggers?}
    %\item \textit{What will the model prediction be if seeing nothing as input?}
    \item \textit{Will the model predict anything even if it sees nothing?}
    \vspace{-0.5em}
\end{itemize}
Inasmuch as we have described the learning process of the model, we know how to obtain model prediction given full context and can easily answer the first question.
The second and third one, however, are hypothetical questions whose answers reflect the confounding biases that we would like to mitigate.
With \emph{attention masks} in recent PLMs \cite{devlin-etal-2019-bert, liu2019roberta, joshi2020spanbert, lan2019albert}, our model can be endowed with \emph{imagination ability} effortlessly.
By inputting a counterfactual instance with the context masked while maintaining the spans of event triggers to the model, we can obtain the model prediction given event trigger names only, which we denote by $\check{\yb}$.
And by sending %inputs with full context masked
an empty (counterfactual) instance, we obtain the model prediction where no textual information is given, which we denote by $\bar{\yb}$.
Intuitively, the two terms $\check{\yb}$ and $\bar{\yb}$ thereof provide measurements for the trigger bias and label bias.

Our goal is to use the biases %distilled from
assessed from model prediction (on counterfactual instances) to generate debiased categorical distribution. We remove the event trigger bias and the frequent label bias via element-wise subtraction, which is proved to be simple yet empirically effective \cite{qian-etal-2021-counterfactual}:
\vspace{-0.5em}
\begin{equation}
    \label{eqn:bias_mitigation}
    \yb' = \yb - \beta_1 \check{\yb} - \beta_2 \bar{\yb},
\end{equation}
where $\yb'$ denotes the debiased categorical distribution and the $\beta$'s are independent parameters for balancing the terms that represent biases.
We find the optimal values for $\beta_1$ and $\beta_2$ on different datasets\footnote{Using the development splits of datasets.} via grid beam search \cite{hokamp-liu-2017-lexically}:
\vspace{-0.25em}
\begin{equation}\label{eq:search}
\resizebox{0.89\hsize}{!}{
    $\hat{\beta_1}, \hat{\beta_2} = \mathop{\arg \max}\limits_{\beta_1, \beta_2}\psi(\beta_1, \beta_2), \ \ \beta_1, \beta_2 \in [a, b]$,
}
\end{equation}

\noindent
where $\psi$ is a metric function (e.g., $F_1$ scores) for evaluation, and $a, b$ are the search boundaries.

In a nutshell, we obtain debiased categorical distribution by removing biases distilled via counterfactual inputs, thus encouraging the model to extract genuinely based on the contextual %description.
content.
Nevertheless, the debiased model is not yet perfect.
A minor drawback lies in that its confidence estimates might have been shifted by element-wise subtraction in  \Cref{eqn:bias_mitigation} though it provides predictions with good evaluation results.
Therefore we employ temperature scaling as our last step to allow for recalibrated confidence measure of the model.

\subsection{Temperature Scaling and OOD Detection}
\label{sec:ts}
The subtraction operation in \Cref{eqn:bias_mitigation} might result in negative values in $\yb'$. 
To provide a proper estimate of the correctness likelihood, we first normalize the probabilities in $\yb'$, where we replace the negative values with a small positive value and %replace the values greater than 1 with a value that approaches 1
clips the values that are greater than 1:
\begin{equation}
\begin{aligned}
\text{norm}(y'_r) = &\ \Big\{ 
\begin{array}{ll} 
\epsilon, & if\ y'_r < 0\\
1-\epsilon,  & if\ y'_r > 1\\
y'_r, & \text{otherwise}
\end{array}~\label{eqn:norm} 
\end{aligned}
\end{equation}
where $\epsilon$ denotes a small positive number, $r \in \mathcal{R}$.
And then we use the inverse function of softmax to obtain the debiased logits vector $\zb'$:
\begin{equation}\label{eq:inverse_softmax}
    \zb' = \sigma^{-1}(\text{norm}(\yb')),
\end{equation}
In this way we are able to apply temperature scaling \cite{pmlr-v70-guo17a} over $\zb'$ and get the recalibrated and debiased categorical distribution $\hat{\yb}$: 
\begin{equation}\label{eq:temperature}
    \hat{\yb} = \sigma(\zb'/T),
\end{equation}
where $T>0$ denotes the temperature\footnote{$T$ is obtained by minimizing the negative log likelihood on the dev set. We refer readers to \Cref{sec:nll} for details.}.

To detect OOD inputs, we need to measure the uncertainty of the model predictions.
We use the \emph{entropy} (\Cref{eqn:um-entropy}) of the final categorical distribution $\hat{\yb}$, which captures the uncertainty encapsulated in the entire distribution.
On the dev set with \textsc{Vague} examples, we find the optimal threshold of $\mathcal{H}[\hat{\yb}]$ below which the model predictions are considered equivoques and the inputs are OOD.
%To distinguish between confident and equivocal predictions we obtain the optimal threshold of $\mathcal{H}[\hat{\yb}]$ on the dev set.
\begin{equation}
\label{eqn:um-entropy}
    \mathcal{H}[\hat{\yb}] = -\sum_{k=1}^{|\mathcal{R}|} \hat{y}_k  \ln(\hat{y}_k) \ .
\end{equation}
\noindent
To sum up, we improve the model faithfulness in both training and inference phase with robust event presentations, Dirichlet Prior parameterization, counterfactual analysis and temperature scaling.
The entire workflow is shown in \Cref{fig:model}.

\section{Experiments}
\begin{table*}[!t]
    \centering
    %\resizebox{\columnwidth}{!}{%
    {
    \small
    \begin{tabular}{l|ccc|ccc|ccc}\hline 
    \toprule
    & \multicolumn{3}{c|}{MATRES} &  \multicolumn{3}{c|}{MATRES-DS} & \multicolumn{3}{c}{TDD-man} \\
    Model & micro-$F_1$ & macro-$F_1$ & ECE & micro-$F_1$ & macro-$F_1$ & ECE & micro-$F_1$ & macro-$F_1$ & ECE \\ \hline
   \citet{mathur-etal-2021-timers} & 82.3 & 55.7 & 12.8                   & 76.7 & 52.3 & 16.3 & 82.1 & 52.8 & 20.3 \\
   \citet{trong2022selecting}      & \textbf{83.4} & \textbf{56.4} & 13.0 & 77.9 & 52.7 & 15.4 & 82.7 & 52.3 & 14.2 \\
   \CC{20}Ours                            & \CC{20}82.7 & \CC{20}56.3 & \CC{20}3.4                    & \CC{20}\textbf{78.7} & \CC{20}\textbf{54.7} & \CC{20}4.0  & \CC{20}\textbf{83.1} & \CC{20}52.9 & \CC{20}\textbf{5.8} \\
    \midrule
    Ours w\slash o TI                   & 81.8 & 55.2 & \textbf{2.0}           & 77.3 & 52.4 & 8.6  & 79.5 & \textbf{66.4} & 21.9 \\
    Ours w\slash o DP                   & 81.3 & 55.2 & 11.8                   & 77.5 & 52.0 & 12.9 & 79.3 & 50.5 & 14.5 \\
    Ours w\slash o CA                   & 80.3 & 54.7 & 5.0                    & 78.6 & 52.9 & \textbf{3.4}  & 83.0 & 52.7 & 6.4\\
    Ours w\slash o TS                   & 82.6 & 56.1 & 49.6                   & \textbf{78.7} & \textbf{54.7} & 15.8 & \textbf{83.1} & 52.9 & 31.0 \\
    \bottomrule
    \end{tabular}
    }\vspace{-0.5em}
    \caption{Model performance on MATRES, MATRES-DS, and TDD-man for event \temprel extraction. The results of ablation study are shown in the last four rows, where TI, DP, CA and TS respectively stand for the four components in our model: Tense Information, Dirichlet Prior, Counterfactual Analysis and Temperature Scaling. 
    Note that the numbers we report on MATRES-DS and TDD-man are model performances \emph{under distribution shifts}.
    }
    \label{tab:temprel}
    \vspace{-0.5em}
\end{table*}
\begin{table}[!t]
    \centering
    \setlength{\tabcolsep}{2pt}
    %\resizebox{\columnwidth}{!}{%
    {
    \small
    %\begin{tabular}{p{1.7cm}|p{0.5cm}p{0.5cm}|p{0.55cm}p{0.5cm}|p{0.5cm}p{0.5cm}}\hline 
    \begin{tabular}{l|cc|cc|cc}\hline 
    \toprule
    %& \multicolumn{2}{p{1.0cm}|}{MATRES} &  \multicolumn{2}{p{1.0cm}|}{MATRES-DS} & \multicolumn{2}{p{1.0cm}}{TDD-man} \\
    & \multicolumn{2}{c|}{MATRES} &  \multicolumn{2}{c|}{MATRES-DS} & \multicolumn{2}{c}{TDD-man} \\
    Model & Acc & MED & Acc & MED  & Acc & MED  \\ \hline
   \citet{mathur-etal-2021-timers} & 43.5 & 1.44 & 32.1 & 1.75 & 37.3 & 1.49\\
   \citet{trong2022selecting}      & 44.7 & 1.36 & 28.0 & 1.96 & 30.5 & 1.55\\
   \CC{20}Ours                     & \CC{20}\textbf{48.2} & \CC{20}\textbf{1.28} & \CC{20}\textbf{43.5}                    & \CC{20}\textbf{1.55} & \CC{20}\textbf{51.7} & \CC{20}\textbf{1.06}  \\
    \midrule
    Ours w\slash o TI                   & 45.8 & 1.37 & 34.5 & 1.66 & 27.1 & 1.87  \\
    Ours w\slash o DP                   & 38.7 & 1.48 & 28.6 & 1.93 & 23.3 & 1.85 \\
    Ours w\slash o CA                   & 43.5 & 1.34 & 39.3 & 1.63 & 49.7 & 1.11 \\
    Ours w\slash o TS                   & \textbf{48.2} & \textbf{1.28} & \textbf{43.5} & \textbf{1.55} & \textbf{51.7} & \textbf{1.06}\\
    \bottomrule
    \end{tabular}
    }\vspace{-0.5em}
    \caption{Model performance on MATRES, MATRES-DS, and TDD-man for timeline construction. The metrics are exact match accuracy (Acc) and minimum edit distance (MED) between prediction and ground truth.
    }\label{tab:timeline}
    \vspace{-0.5em}
\end{table}
In this section, we describe the experiments\footnote{We refer readers to \Cref{sec:exp_setup} for the discussion of experimental setup.} %to evaluate model faithfulness 
on two tasks: \temprel extraction and timeline construction. We first introduce the datasets that we adopt or create for evaluation (\Cref{sec:datasets}), followed by the evaluation protocols (\Cref{sec:eval_protocol}). Evaluation results %of our model 
are discussed in \Cref{sec:results} before we provide a detailed ablation study and case study in \Cref{sec:ablation} and \Cref{sec:case}.
\subsection{Datasets}

We evaluate using the following datasets, for which statistics are given in \Cref{sec:stats}.

\label{sec:datasets}
\stitle{MATRES} \cite{ning-etal-2018-multi} is a \temprel benchmark annotated with the multi-axis scheme that helps  achieve higher inter-annotator agreements (IAA) than previous benchmark datasets \cite{cassidy-etal-2014-annotation, styler2014temporal, ogorman-etal-2016-richer}.
%MATRES is composed of 275 news documents and the train/dev/test split is 183/72/20 documents.
Four relations are annotated for the start time comparison of event pairs in 275 documents, namely \textsc{Before}, \textsc{After}, \textsc{Simutaneous}, and \textsc{Vague}.
We train our model on the training set of MATRES, and evaluate our model on the dev and test sets of MATRES, MATRES-DS and TDDiscourse, which we introduce next.

\stitle{MATRES-DS} is an evaluation dataset that we created %to serve as an  set 
with distribution shifts (DS) compared to MATRES. 
Since one of our goals is to mitigate the bias of event triggers in the training data, we examine whether our proposed model stays uninfluenced %from the bias by altering 
when the distribution of event triggers is altered.
We replace frequent triggers in the MATRES dev and test sets that appear within the top 5K frequent lemmas\footnote{\url{https://www.wordfrequency.info/}} with their uncommon synonyms, and replace infrequent triggers with their frequent synonyms from the list of frequent lemmas.
MATRES-DS also presents a mismatch between the training and test distributions, or dataset shift \cite{quinonero2008dataset}, where distributional uncertainty often arises.

\stitle{TDDiscourse} \cite{naik-etal-2019-tddiscourse} is a dataset for discourse-level event temporal ordering, in which \temprel's between global long-distance event pairs are annotated.
As another data source with distribution shifts 
compared to MATRES, we adopt the manually annotated subset of TDDiscourse, namely TDD-man, in our experiments.
%, in which 4,000/650/1,500 \temprel's are annotated in the train/dev/test sets.
The \temprel set $\mathcal{R}_{T}$\footnote{$\mathcal{R}_{T}$ = \{\textsc{Before}, \textsc{After}, \textsc{Simultaneous}, \textsc{Includes}, \textsc{Is Included}\}.} annotated in TDDiscourse is a superset of the \temprel set $\mathcal{R}_{M}$ defined in MATRES.
Given that TDD-man serves as evaluation data on which we do not train our model, a relation in $\mathcal{R}_{M} \cup \{\textsc{Vague}\}$ is predicted for each pair of events in the test set of TDD-man.

\subsection{Evaluation Protocols}
\label{sec:eval_protocol}
For \textbf{event \temprel extraction}, we compare our model with the current and previous SOTA models \cite{trong2022selecting, mathur-etal-2021-timers} trained on MATRES.
The models are evaluated on not only how precise and selective their extraction is on ID data (MATRES), but are also examined for their generalizability under distribution shifts (MATRES-DS and TDD-man).
We report micro-$F_1$ score as an evaluation metric following previous papers.
We also report macro-$F_1$, which %is better %suited to 
reflects the %biases' severity
fairness of model prediction,
and expected calibration error (ECE) %\footnote{We refer reader to \Cref{sec:appendix} for a detailed definition of ECE.} 
that approximates the \emph{difference in expectation between confidence and accuracy}. 
%We refer readers to \Cref{sec:ece} for a 
The definition of ECE is provided in \Cref{sec:ece}.

We also apply our model to the \textbf{timeline construction} task, where the goal is to sort a list of events in a document in chronological order.
%The method of timeline construction is as follows
%in the following process: a) the model first predicts for all possible event pairs in the given list; b) and then a directed graph $G$ is built with events as vertices and non-\textsc{Vague} \temprel's as edges; c) next we remove the edges in $G$ with lowest confidence scores until $G$ becomes a directed acyclic graph (DAG); d) and finally the linear ordering of the vertices in $G$ generated by topological sorting is considered as the predicted timeline.
To construct the timeline, the model first constructs a directed graph $G$ with predicted non-\textsc{Vague} \temprel's between every event pairs. 
Then, edges in $G$ with lowest confidence scores are removed until $G$ becomes a directed acyclic graph (DAG).
Finally, the timeline is generated as the linear ordering of the vertices in the DAG by topological sorting. 
In this way, we circumvent the possible conflicts in model predictions for timeline construction and the \emph{faithful removal} of least confident edges serves as an examination on the quality of model-predicted confidence.
On the three datasets, we report the accuracy of exact match and the average minimum edit distance between predicted and ground truth timelines as evaluation metrics.
\subsection{Results}
\label{sec:results}
\begin{figure}[t]
    \centering
    \includegraphics[width=\linewidth]{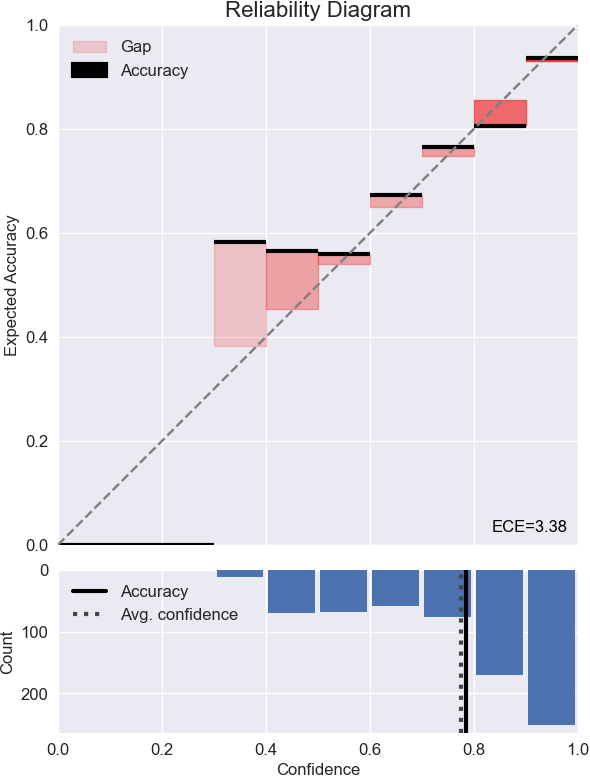}
    \vspace{-1em}
    \caption{Reliability diagram and confidence histogram of our model predictions on test set of MATRES.}\label{fig:diagram}
    \vspace{-1em}
\end{figure}

In \Cref{tab:temprel}, we report the \temprel extraction results. On MATRES, the SOTA model \cite{trong2022selecting} still offers the best performance in terms of micro-$F_1$ whereas our model achieves comparable macro-$F_1$ score and lower calibration error.
In contrast, our proposed \emph{faithful} \temprel extractor outperforms baseline methods in terms of all evaluation metrics under the dataset shifts caused by replacement of event triggers in MATRES-DS and longer context distances between global event pairs in TDD-man.
Specifically, our model shows a significant gain of 2.0\% macro-$F_1$ and 0.8\% micro-$F_1$ over the SOTA model on MATRES-DS and surpasses the previous SOTA model on TDD-man by 1.0\% micro $F_1$, not to mention the improvements on %calibration performances.
confidence calibration.
We attribute this superior performance under dataset shifts to the mitigation of biases from prior knowledge and training set statistics as well as the techniques we employ to improve predictive uncertainty estimation.
For a visual illustration of model calibration, we present the reliability diagram that plots the expected sample accuracy as a function of confidence in \Cref{fig:diagram}. 

\Cref{tab:timeline} exhibits similar observations: our model outperforms both baselines on timeline construction by a large margin in terms of both metrics.
Specifically, under dataset shifts within MATRES-DS and TDD-man, our model surpasses the best baseline by 11.4\% and 14.4\% in accuracy, while drastically reducing the minimum edit distance by relatively 11.4\% and 28.9\%.
Evidently, the capabilities of selective prediction and bias mitigation make our model stand out in complex scenarios like timeline construction, whereas the bias and inferior calibration of existing models exacerbate unfaithful extractions when multiple decisions have to be made simultaneously. % and the model has to reject unconfident predictions according to the predicted categorical distributions.
\subsection{Ablation Study}
\label{sec:ablation}
To analyze the effect of each model component, we conduct an ablation study of our model where we remove one component at a time (see \Cref{tab:temprel})\footnote{We leverage cross-entropy as the training loss when we remove Dirichlet Prior in the training phase.}.
%And given that the removal of counterfactual analysis does not shift the categorical distribution, the temperature scaling part is also removed.
We observe on MATRES that, without the counterfactual analysis component, the model performance becomes worse by 2.4\% in micro-$F_1$ and 1.6\% in macro-$F_1$.
Under dataset shifts, the model performance is reduced by 3.8\% in micro-$F_1$ and 2.4\% in macro-$F_1$ on TDD-man without the parameterization of Dirichlet Prior.
The model performance in terms of $F_1$ scores is slightly influenced by taking away the temperature scaling component while the model calibration severely degrades.

From the ablation results in \Cref{tab:timeline}, we notice that temperature scaling has modest effects on the model performances, while Dirichlet Prior plays the most important role towards faithful timeline construction.
It is also noteworthy that tense information considerably benefits the model to generalize well under distribution shifts in that it provides a useful feature applicable to all domains.

% Because the bias in MATRES / MATRES-DS is more severe, so the performance difference is larger
% Because TDD includes global event pairs which is more comprehensive, the bias is less severe in the dataset.

%1. Affixing tense information into event representations not only improves the model performance on MATRES, but also yields better results on shifted-distribution and OOD data.
%2. Dirichlet Prior brings about better 

%\subsection{Ablation Study}
%\todo{The model trained with cross-entropy loss are worse calibrated. It does not cooperate well with bias mitigation since after subtracting biases, the confidence is hard to reflect the true correctness likelihood, even with  temperature scaling afterwards.}

\subsection{Case Study}
\label{sec:case}
As shown in \Cref{fig:timeline}, we provide a case study on timeline construction for three events. %are investigated. % and the prediction results and confidences are shown in the table.
%as follows: (\textcolor{blue}{\textsc{slayings}}, \textcolor{blue}{\textsc{careful}}): \textsc{Before}, 0.92; (\textcolor{blue}{\textsc{careful}}, \textcolor{blue}{\textsc{seen}}): \textsc{Before}, 0.72; (\textcolor{blue}{\textsc{slayings}}, \textcolor{blue}{\textsc{seen}}): \textsc{After}, 0.51. 
The reason why our model predicts \textsc{After} for the third pair is probably due to the misleading temporal cues in text, \textcolor{olive}{Thursday} and \textcolor{olive}{Tuesday}, while the long distance between events undermines the confidence for this prediction.
When our model builds a directed graph with three relations, a cycle is identified and the edge with lowest confidence is removed from the graph, and thus our model constructs the correct timeline.
Without tense information, the model makes wrong prediction concerning the second event whose trigger is an adjective.
And without Dirichlet Prior or temperature scaling, %it can be noticed that the model calibration becomes worse.
the model calibration becomes noticeably worse.

\begin{figure}[t]
\begin{minipage}{\linewidth}
\noindent
\centering
\fbox{%
    \parbox{0.9\linewidth}{
    \fontsize{10pt}{13pt}\selectfont
    A new Essex County task force began delving \textcolor{olive}{Thursday} into the \textcolor{blue}{\textsc{$e_1$:slayings}} of 14 people ... 
    officials have been \textcolor{blue}{\textsc{$e_2$:careful}} not to draw any firm conclusions, leaving open the possibility of a serial killer ... 
    %There have been no arrests in any of the slayings. 
    ``I haven't \textcolor{blue}{\textsc{$e_3$:seen}} a pattern yet,'' said Patricia Hurt, the Essex County prosecutor, who created the task force on \textcolor{olive}{Tuesday}.
    }
}

\vspace{0.4em}
    \centering
    \setlength{\tabcolsep}{3pt}
    {%
    \small
    \begin{tabular}{l|ccc|c}
    \hline
    \toprule
     Model & $(e_1, e_2)$ & $(e_2, e_3)$ & $(e_1, e_3)$ & Timeline \\ \hline
     Gold & \textsc{b} & \textsc{b} & \textsc{b} & $(e_1, e_2, e_3)$ \\
     Ours & \textsc{b}, 0.92 & \textsc{b}, 0.72 & \CC{18}\textsc{a}, 0.51 & $(e_1, e_2, e_3)$ \\ \hline
     Ours w\slash o TI & \textsc{b}, 0.99 & \RED{18}\textsc{a}, 0.53 & \textsc{b}, 0.63 & \RED{18}$(e_1, e_3, e_2)$ \\
     Ours w\slash o DP & \textsc{b}, 0.92 & \RED{18}\textsc{a}, 0.34 & \textsc{b}, 0.52 & \RED{18}$(e_1, e_3, e_2)$ \\
     Ours w\slash o CA & \textsc{b}, 0.94 & \textsc{b}, 0.43 & \textsc{b}, 0.49 & $(e_1, e_2, e_3)$ \\
     Ours w\slash o TS & \textsc{b}, 0.43 & \textsc{b}, 0.38 & \CC{18}\textsc{a}, 0.36 & $(e_1, e_2, e_3)$ \\
             % \hline
    \bottomrule
    \end{tabular}
    }
    \vspace{-0.5em}
    \caption{Case study on timeline construction for one of the documents in TDD-man. The table shows predicted \temprel's and confidence for three event pairs, where \textsc{b} stands for \textsc{Before} and \textsc{a} stands for \textsc{After}. The cells in \colorbox{red!18}{light red} and \colorbox{blue!18}{light blue} are wrong predictions and relations removed in timeline construction, respectively.}\label{fig:timeline}
    \vspace{-1em}
\end{minipage}
\end{figure}

\section{Conclusion}
\label{sec:conclusion}
In this paper, we investigate on improving faithfulness for event \temprel extraction from two perspectives.
To enhance the selectiveness of model predictions, we parameterize a Dirichlet Prior over the model-predicted categorical distribution to regularize the model to behave differently with ID and OOD data.
%In the training phase, we add tense information to event representations and  %, thus enhancing the selectiveness of model predictions. 
To mitigate two types of biases from PLMs and training data, we add tense information to obtain robust event representations and conduct a counterfactual analysis to reduce the risk of carrying prediction shortcuts into inference.
We also employ temperature scaling to combine the two faithful perspectives, which recalibrates the confidence measure of the model after bias mitigation.
Through experimental analysis on MATRES, MATRES-DS, and TDDiscourse, we demonstrate that our model faithfully extracts event temporal relations and timelines from text, %especially 
so as to generalize well under distribution shifts.

\section*{Limitations}
%We would like to caution about shortcomings of the proposed system in terms of misclassifications on event pairs requiring real-world commonsense reasoning and domain shift, given that the articles used for training purposes are mainly excerpts from previous news reports.
As the event representation introduced in our method is augmented with tense information, it potentially leads to limitations when applying to languages other than English, especially tenseless languages and languages having fewer tenses.
The training of our models also requires considerable GPU resources which might produce environmental impacts, though the inference stage does not take up much computational resources.

\section*{Ethics Statement}
There are no direct societal implications of this work. The proposed method attempts to provide high-quality and faithful event \temprel extraction and timeline construction. We believe that the intellectual merits of developing robust event-centric information extraction methods are demonstrated by this work.
For any information extraction methods, real-world open source articles used to extract information may contain societal biases.
Extracting event-event relations from articles with such biases may spread the bias into the acquired knowledge.
Yet we believe that the proposed method can benefit various downstream NLP/NLU tasks like event prediction, task-oriented dialogue systems and risk detection.

% Entries for the entire Anthology, followed by custom entries
\bibliography{anthology,custom}
\bibliographystyle{acl_natbib}
\clearpage
\appendix
\input{appendix}

\end{document}

%% file: appendix.tex
\section{Appendix}
\label{sec:appendix}
\subsection{Example of event context affixed tense information}
\label{sec:tense_eg}
\begin{figure}[h]
\begin{minipage}{\linewidth}
\noindent
\fbox{%
    \parbox{0.95\linewidth}{
    \fontsize{11pt}{13pt}\selectfont
    Original context: 
    [CLS] For his part, Fidel Castro is the ultimate political survivor.[SEP] People have \textcolor{blue}{\textsc{predicted}} his demise so many times, and the US has \textcolor{blue}{\textsc{tried}} to hasten it on several occasions.[SEP] Time and again, he endures.[SEP] \\ 
    
    Context with affixed tense information:
    [CLS] For his part, Fidel Castro is the ultimate political survivor.[SEP] People have @ * \textsc{\textit{Present Perfect Simple}} * \textcolor{blue}{\textsc{predicted}} @ his demise so many times, and the US has \# $\wedge$ \textsc{\textit{Present Perfect Simple}} $\wedge$ \textcolor{blue}{\textsc{tried}} \# to hasten it on several occasions.[SEP] Time and again, he endures.[SEP] 
    }
}
    \caption{An example of the original context of event pair (\textcolor{blue}{\textsc{predicted}}, \textcolor{blue}{\textsc{tried}}) and the context after affixing tense information to corresponding event spans.}
    \label{fig:affix}
\end{minipage}
\end{figure}

\subsection{Definition of ECE}
\label{sec:ece}
Expected Calibration Error (ECE) metric \cite{pmlr-v70-guo17a} measures exactly the difference in expectation between confidence and accuracy. Empirically it is approximated by dividing the data into $M$ confidence based bins, i.e., $B_m$ (where $m \in
\{1, 2, ..., M\}$) contains all datapoints $i$ for which predicted confidence $p_i$ lies in ($\frac{m-1}{M}, \frac{m}{M}$]. If acc($B_m$) and conf($B_m$) denotes the average accuracy and prediction confidence for the points in $B_m$, ECE is defined as:
\begin{equation}
\label{eqn:ece}
\text{ECE} = \sum_{m=1}^{M} \frac{|B_m|}{n} \bigg|\acc(B_m) - \conf(B_m)\bigg|,
\end{equation}
where $n$ is the number of samples.
The difference between acc and conf for a given bin represents the calibration gap (red bars in reliability diagrams – e.g. \Cref{fig:diagram}). We use ECE as the primary empirical metric to measure model calibration. 

\subsection{Negative Log Likelihood}
\label{sec:nll}
Negative log likelihood is a standard measure of a probabilistic model's quality. It is also referred to as the cross entropy loss in the context of deep learning. Given a probabilistic model $\hat{\pi}(Y|X)$ and $n$ samples, NLL is defined as:
\begin{equation}
  \scrL = -\sum_{i=1}^{n}\log(\hat{\pi}(y_i|\xb_i))
\end{equation}
It is a standard result that, in expectation, NLL is minimized if and only if $\hat{\pi}(Y|X)$ recovers the ground truth conditional distribution $\pi(Y|X)$.
The temperature $T$ in temperature scaling is optimized with respect to NLL on the dev sets. 

\subsection{Dataset Statistics}
\label{sec:stats}
MATRES is composed of 275 news documents and the train/dev/test split is 183/72/20 documents where 6336/6404/818 event pairs are annotated respectively.
The same statistics hold for MATRES-DS since we only change the event triggers in the inputs instead of the labels.
In TDDiscourse,
4,000/650/1,500 and 32609/1435/4258 \temprel's are annotated in the train/dev/test sets of TDD-man and TDD-Auto, respectively.
%We fine-tune the hyper-parameters in our model using the development set of
%the MATRES dataset.

\subsection{Experimental Setup and Hyperparameter Setting}
\label{sec:exp_setup}
In the training phase, we fine-tune the pre-trained 1024-dimensional Big Bird \cite{zaheer2020big} to encode the context of event triggers.
We obtain the tense information of event triggers with an off-the-shelf tense identifier\footnote{\url{https://tense-sense-identifier.herokuapp.com/}}.
The parameters of the model are optimized using AMSGrad \cite{reddi2018convergence} with the learning rate set to $\num{5e-6}$, batch size set to 20, and the training process is limited to 40 epochs on a server with Nvidia A6000 GPU.
All experiments are repeated with five different random seeds and the results reported are their average.
To obtain $\bm{\alpha}_{s}$ in \Cref{eqn:id_loss},
we smooth the target means to redistribute a small amount of probability density to the other corners of the Dirichlet.
In our experiments, we set $\lambda_1 = \lambda_2 = 1$ in \Cref{eqn:loss}.
On the dev set of TDD-man the optimal $\beta$'s of the model in \Cref{eqn:bias_mitigation} are $\beta_1 = -0.4, \beta_2 = 0.6$, where the search bounds, $a$ and $b$ equal to -1 and 1.
%This suggests that the event trigger bias learned in the training set of MATRES is strengthened and the label bias is weakened.

%\subsection{Hyperparameters}